\documentclass[11pt]{article}
\usepackage[margin=1in]{geometry}
\usepackage[T1]{fontenc}
\usepackage{lmodern}
\usepackage{microtype}
\usepackage{amsmath,amssymb,amsthm}
\usepackage{booktabs}
\usepackage{graphicx}
\usepackage[font=small,labelfont=bf]{caption}
\usepackage{enumitem}
\usepackage[numbers,sort&compress]{natbib}
\usepackage{xcolor}
\usepackage[colorlinks=true,linkcolor=blue!50!black,citecolor=blue!50!black,urlcolor=blue!50!black]{hyperref}
\usepackage[capitalise,noabbrev]{cleveref}

\newtheorem{proposition}{Proposition}
\newtheorem{corollary}{Corollary}
\theoremstyle{definition}
\newtheorem{definition}{Definition}
\theoremstyle{remark}

\newcommand{\E}{\mathbb{E}}

\newcommand{\Cov}{\operatorname{Cov}}
\newcommand{\aab}{\texorpdfstring{$\alpha'\beta'$}{alpha'beta'}}

\title{\textbf{When Does the Real Wiring Win?}\\[2pt]
\large Testing connectome structure--function claims against unmeasured physiology}
\author{Yeyuan Liu\\
\small University of Virginia\\
\small \texttt{pka5qg@virginia.edu}}
\date{}

\begin{document}
\maketitle

\begin{abstract}
Connectome-constrained models are often used to argue that a measured wiring diagram outperforms shuffled or rewired wiring, and hence that structure shapes function. Such comparisons can confound wiring with how the models are tuned: a connectome fixes who connects to whom, not the excitability, gains or plasticity rates the model also needs, and modellers often tune these until the real-wiring model works. We separate three estimands: physiology tuned on the real wiring and transferred to nulls (P1), each wiring tuned on its own (P2), and P2 restricted to physiology consistent with independent data (P3). The biases involved are known in machine learning (under-tuned baselines, inherited weights, winner's curse); we measure what they do to biological conclusions, using synthetic ground truth, placebo nulls and three \textit{Drosophila} circuit models. Some claims survive every protocol we test: segregated reward and punishment dopamine improves learning, and head-direction topography is required for a persistent bump. Others do not. In the mushroom body, the real Kenyon-cell output wiring beats a cross-subtype identity shuffle by $0.09$ under P1 and P2 but loses to it by $0.09$ under P3 (95\% CIs exclude zero; held-out runs confirm). The verdict is set by the share of \aab{} cells among active Kenyon cells, which P1 selects low; at the share implied by imaging (about 44\%) the real-wiring model fails to learn, exposing a model inadequacy rather than a deficit of the fly's wiring. In flyvis, P1 penalises a mirrored connectome, which has a constructed solution within 0.07 (median) of the real error, more than a spatially shuffled one in 50 of 50 members; the ranking survives ten-fold longer retraining but depends on the checkpoint at fifty-fold. A credible structural claim needs matched optimisation, data-constrained physiology and a model that still works there.
\end{abstract}

\section{Introduction}
\textbf{A structural claim is a claim about a comparison, not only about a score.} Synapse-resolution connectomes of the adult fly brain \citep{dorkenwald2024,schlegel2024} have made connectome-constrained models of function practical \citep{lappalainen2024,shiu2024}, and with them a new kind of argument. Recent studies compare models built on the measured wiring with the same models built on shuffled or rewired wiring \citep{cho2026,park2026,zhou2026}. When the measured wiring performs better, the wiring is taken to matter for the function, but the conclusion depends on what the controls preserve and how the models are evaluated \citep{park2026}.

\textbf{The comparison has a gap.} A connectome fixes the graph and synapse counts. It does not fix how excitable each cell type is, how strong a synapse is per contact, how fast cells integrate, or how fast plasticity runs. These quantities can make very different circuits behave alike, and very similar circuits behave differently \citep{prinz2004,marder2006}; connectome constraints alone underdetermine dynamics \citep{beiran2024}. Modellers therefore choose them, usually once, and often by adjusting them until the real-connectome model performs the task. Null wirings then inherit parameters selected for the real one.

\begin{figure}[t]
\centering
\includegraphics[width=\linewidth]{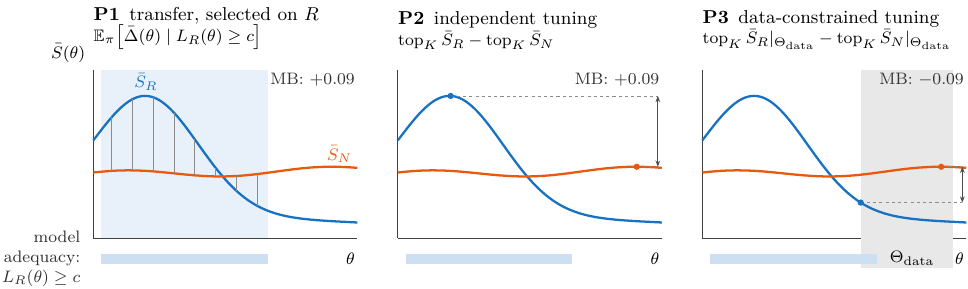}
\caption{\textbf{Three protocols answer different questions, and all depend on model adequacy.} Schematic for one unmeasured parameter $\theta$; the curves are illustrative, not data. Blue: seed-averaged score of the real wiring, $\bar S_R(\theta)$; orange: a null, $\bar S_N(\theta)$. P1 averages the paired difference (grey lines) over physiology selected because the real-wiring model meets the behavioural criterion, $L_R(\theta)\ge c$ (blue shading), a selection that tends to favour the real wiring (\cref{sec:theory}). P2 compares the best case of each wiring (dots; in practice the mean of the $K$ best parameter sets). P3 does the same within the physiology allowed by independent data, $\Theta_{\mathrm{data}}$ (grey). Bars: physiology where the real-wiring model meets the criterion. A comparison made where it fails, as in most of $\Theta_{\mathrm{data}}$ here, tests the model rather than the wiring. MB: mushroom-body estimates for the KC$\to$MBON wiring against the identity shuffle (\cref{sec:mb}).}
\label{fig:schematic}
\end{figure}

\textbf{Different comparisons answer different questions} (\cref{fig:schematic}). Tuning physiology until the real wiring works and transferring it to the nulls (P1) asks whether physiology tuned for one wiring fits another, and it rewards the wiring it was tuned on. Tuning each wiring on its own (P2) removes this asymmetry, but its best case may need physiology the animal does not have. Restricting the search to physiology allowed by independent data (P3) asks the biological question, provided the model still performs the task there. None of these problems can be seen at a single parameter setting, and selecting parameters on the real model's success conditions the comparison on that success. In neuroscience this is a familiar trap in data analysis \citep{kriegeskorte2009}, and machine learning knows it as under-tuned baselines, weight inheritance and the winner's curse (\cref{sec:related}). Our aim is not to discover the bias but to measure what it does to biological conclusions drawn from connectomes, where the shared parameters are unmeasured physiology and a fair comparison must also respect what experiments constrain.

\textbf{Parameter choices can reverse a structural claim.} In a mushroom-body model, the real Kenyon-cell output wiring beats cross-subtype nulls under P1 and also under P2, where each wiring is tuned on its own, the usual remedy in machine learning; it loses only under P3, and at the subtype composition implied by imaging the real-wiring model fails to learn at all. In the trained flyvis visual system, P1 penalises a mirrored connectome with a known near-real solution more than a spatially shuffled one in every one of 50 pretrained members. In a head-direction ring attractor, by contrast, the real topology is needed for a persistent bump under every protocol we test.

\paragraph{Contributions.}
\begin{enumerate}[leftmargin=*,itemsep=2pt]
\item \emph{A framework for comparing connectome-constrained models.} We separate parameter transfer, independent tuning and data-constrained tuning (P1--P3, \cref{sec:framework}), state the selection and inherited-weight biases they are exposed to with closed forms for their size (\cref{prop:gauss,prop:warm}), and check these against synthetic ground truth and placebo nulls.
\item \emph{A cross-circuit test of structural claims.} Across associative learning, attractor memory and motion vision, we separate conclusions that survive a change of protocol from those that reverse, and confirm the main reversal on held-out samples, model variants and spatially constrained nulls.
\item \emph{A link from protocol sensitivity to measurement and model adequacy.} We identify the one unmeasured quantity that decides the fragile claim (the subtype composition of odour-responsive Kenyon cells), use published imaging to estimate it, and show that at that value the model itself fails, which no comparison between wirings can reveal on its own.
\end{enumerate}

\section{Related work}\label{sec:related}
\paragraph{Connectome-constrained models of function.} Lappalainen et al.\ \citep{lappalainen2024} train ensembles of 50 visual-system models that differ in which connectivity information is given and which is learned, and evaluate them mainly by how well they predict measured neural responses. Their ensembles already address non-uniqueness of trained parameters; our question is whether that uncertainty changes verdicts about wiring. Two of their observations bear on ours: models with random network parameters but trained decoders perform consistently worse on the task, and two ensemble clusters with opposite T4c preferred directions reach nearly identical average task errors (5.297 and 5.316), so a task score alone does not identify mechanism. Whole-brain models compare the real connectome against shuffled weights that keep the global weight distribution \citep{shiu2024} or against degree-preserving rewired graphs \citep{cho2026}; an embodied brain-and-nerve-cord model keeps its published parameters for both the connectome and a degree- and sign-preserving rewiring \citep{minnick2026}. Evolved agents have been compared against rewired graphs \citep{park2026}: Park shows that what the null preserves at the sensory--motor boundary changes the verdict; there each wiring is evolved separately after matching mean activity, which corresponds to our P2, and physiology is not swept. Zhou and Hasler \citep{zhou2026} compare representational geometry of the flyvis ensemble against weight shuffles. FlyHack \citep{dhawan2026} shows that mechanistic conclusions in flyvis depend on the trained solution and the intervention. Duan, Dong and Fiete \citep{duan2025} derive ring-attractor dynamics from head-direction connectivity. Yad and Nabila \citep{yad2026} show that differences between reconstructions of the fly compass circuit can rival differences between individuals and can break trained circuit models; this concerns uncertainty in the wiring itself, whereas we hold the wiring fixed and vary what it leaves unmeasured. None of these varies unmeasured physiology to test whether a real-versus-null verdict survives, or treats the comparison protocol as a source of bias. In macroscale network neuroscience, structure--function coupling and spatially constrained nulls are established tools \citep{suarez2020,vasa2022}; those analyses relate structural and functional maps or correlations, whereas we compare mechanistic models whose unmeasured quantities are physiological parameters. In concurrent work on human connectomes used as fixed reservoirs, Bertamini et al.\ \citep{bertamini2026} separate the effect of which regions are connected from that of which connection carries which weight; weight placement accounts for 28--75\% of the reported adult advantage, depending on the null, while newborn connectomes keep a wiring advantage that a matched small-world graph also reaches. Their confound lies in how a null places weights, ours in how physiology is chosen for the wirings being compared; the two corrections are complementary.

\paragraph{Closest prior work.} In a preprint, Dhiman \citep{dhiman2026} reports in flyvis, on MovingEdge direction decoding, that an early-learning advantage of the connectome over a naive random graph disappears when checkpoint initialisation is replaced by shared random initialisation and the null preserves degree sequence; its conclusions are explicitly limited to 5 and 10 optimisation steps. We share its observation that a connectome-trained initialisation favours the connectome. We differ in varying unmeasured physiology systematically, in showing that some claims survive every protocol we test while others reverse, in measuring the bias against a closed form, and in using an exact oracle that bounds a placebo wiring's attainable error independently of training length.

\paragraph{Analogues in machine learning.} The underlying problem is familiar in machine learning. Baselines given less tuning than the proposed method look worse than they are \citep{melis2018,lucic2018,musgrave2020}, and benchmark verdicts depend on data sampling, initialisation and hyperparameter choice \citep{bouthillier2021}. Selecting the best of many configurations and reporting its score is a winner's-curse problem \citep{zrnic2024}; our \cref{prop:gauss} is an instance of it. Pruned architectures were credited with value that disappears when they are trained from scratch rather than fine-tuned from the dense network's weights \citep{liu2019rethinking}, and whether a sparse structure helps depends on the initialisation it is paired with \citep{frankle2019lottery}. In weight-sharing architecture search, candidates scored with inherited weights are ranked poorly relative to stand-alone training \citep{yu2020nas}, and weight sharing biases rankings consistently \citep{zhang2020ws}; our flyvis warm-start result is the same effect. P1 and warm-start retraining are the connectome versions of under-tuned baselines and inherited weights. What differs here is that the shared quantities are biological parameters that have not been measured, so a fair comparison needs not only equal tuning (P2) but also a restriction to physiology consistent with data (P3), and that the null wirings are chosen to test specific anatomical claims.

\paragraph{Parameter degeneracy and null models.} Circuit models are degenerate \citep{prinz2004,marder2006} and sloppy, with a few stiff and many soft parameter directions \citep{gutenkunst2007}. Null models are central to network neuroscience \citep{vasa2022}; we use degree-preserving double-edge swaps \citep{maslov2002}. Selection on an outcome biases estimates in econometrics \citep{heckman1979} and in neuroimaging \citep{kriegeskorte2009}; our Propositions~\ref{prop:exch}--\ref{prop:gauss} adapt this to model comparison.

\section{Framework}\label{sec:framework}
A claim has the form ``real wiring $R$ beats null family $N$ on score $S$''. Let $\theta\in\Theta$ collect unmeasured physiological parameters, drawn from a distribution $\pi$ over plausible ranges. For each $\theta$ and seed $s$ (shared stimuli and noise across wirings) we compute $S_R(\theta,s)$ and $S_N(\theta,s)$, and the paired difference $\Delta_s(\theta)=S_R(\theta,s)-S_N(\theta,s)$.

\paragraph{Per-$\theta$ verdict.} With $n$ seeds, the verdict is ``real better'' if $\bar\Delta(\theta)>\delta$ and at least $n-1$ seeds are positive, ``null better'' in the mirror case, and ``no clear difference'' otherwise. We use $\delta=0.03$ for mushroom-body retention accuracy and $\delta=0.05$ for head-direction scores. We report the fraction of $\theta$ in each class, over all $\theta$ and over \emph{plausible} $\theta$ (those where the real model meets a behavioural criterion $L_R(\theta)\ge c$).

\begin{definition}[Comparison protocols]\label{def:protocols} With $\bar S_G(\theta)$ the seed-averaged score,
\begin{align*}
\text{P1 (shared, tuned until real works):}\quad & \E_\pi\big[\bar\Delta(\theta)\mid L_R(\theta)\ge c\big],\\
\text{P2 (each wiring tuned on its own):}\quad & \textstyle \mathrm{top}_K\,\bar S_R-\mathrm{top}_K\,\bar S_N,\\
\text{P3 (P2 within data-consistent physiology):}\quad & \textstyle \mathrm{top}_K\,\bar S_R\big|_{\Theta_{\mathrm{data}}}-\mathrm{top}_K\,\bar S_N\big|_{\Theta_{\mathrm{data}}},
\end{align*}
where $\mathrm{top}_K$ is the mean of the $K=5$ best parameter sets and $\Theta_{\mathrm{data}}\subset\Theta$ is the region allowed by independent physiological data.
\end{definition}
P1 is current practice. It answers what happens when physiology tuned for the real wiring is used with another wiring; it does not measure which wiring can do better. P2 asks which wiring has the better best case. P3 asks the same within physiology the data allow. P3 requires an independent constraint and is therefore available only where one exists.

\paragraph{Sensitivity.} We report the Spearman correlation between each parameter and $\bar\Delta(\theta)$, and Sobol first- and total-order indices \citep{sobol2001} computed on a gradient-boosted surrogate of $\bar\Delta(\theta)$ fitted to the sweep (cross-validated $R^2$ reported). The parameter with the largest index is the claim's \emph{deciding parameter}.

\paragraph{Placebo nulls.} We label one instance of a null family ``real'' and compare it with another instance of the same family. Any difference is then due to the protocol alone.

\section{Theory}\label{sec:theory}
\begin{proposition}[P1 favours the real wiring under no effect]\label{prop:exch}
Let $(S_R,S_N)$ be exchangeable under $\theta\sim\pi$, so wiring has no effect in distribution. Then for any threshold $c$ with $\Pr(S_R\ge c)>0$,
\[
\E[S_R-S_N\mid S_R\ge c]=\frac{\tfrac12\,\E\big[(S_R-S_N)(\mathbf 1\{S_R\ge c\}-\mathbf 1\{S_N\ge c\})\big]}{\Pr(S_R\ge c)}\;\ge\;0,
\]
with equality only if $S_R$ and $S_N$ fall on opposite sides of $c$ with probability zero.
\end{proposition}
Both factors in the numerator always share a sign because the indicator is monotone (proof in Appendix~\ref{app:proofs}). Selection in practice is often on a behavioural criterion $L_R$ rather than on the score itself. A Gaussian approximation then gives the size of the bias.

\begin{proposition}[Closed-form bias]\label{prop:gauss}
Let $X=S_R-S_N$ and $L_R$ be jointly Gaussian, with $\E X=\mu$ the true effect. Let $p=\Pr(L_R\ge c)$, $\alpha=\Phi^{-1}(1-p)$ and $\lambda(\alpha)=\phi(\alpha)/p$, the inverse Mills ratio. Then
\[
\E[X\mid L_R\ge c]=\mu+\frac{\Cov(X,L_R)}{\sigma_L}\,\lambda(\alpha).
\]
If selection is on the score itself ($L_R=S_R$) and $(S_R,S_N)$ are exchangeable with variance $\sigma^2$ and correlation $\rho$, the bias is $\sigma(1-\rho)\lambda(\alpha)$.
\end{proposition}
The sign of the bias is the sign of $\Cov(X,L_R)$: selection on a criterion favours the real wiring only when the criterion tracks the real model's score more than the null's. \Cref{prop:exch} covers the special case of selection on the score itself; for selection on a separate criterion the direction is an empirical question, which the placebo nulls answer for each family (\cref{sec:placebo}; all six measured biases are positive). The bias grows as the criterion becomes more selective ($\lambda$ increases as $p$ falls), as scores spread more across $\theta$ and seeds ($\sigma$), and as the two wirings respond less alike ($1-\rho$). P2 has no such bias: under exchangeability the best-of-$K$ statistics of the two wirings are identically distributed, so $\E[\text{P2}]=0$.

Gradient-trained models raise a related problem. A natural control retrains every wiring for a fixed budget, starting from a solution trained on the real wiring; Dhiman \citep{dhiman2026} identifies this checkpoint-initialised comparison as biased and replaces it with shared random initialisation. We characterise the bias.
\begin{proposition}[Finite-budget retraining from a real-trained solution]\label{prop:warm}
Near their optima, let each wiring's loss be quadratic, $\mathcal L_G(\varphi)=\mathcal L_G^\ast+\tfrac12(\varphi-\varphi_G^\ast)^\top H_G(\varphi-\varphi_G^\ast)$ with $H_G\succeq0$. Run gradient descent with step $0<\eta\le1/\lambda_{\max}(H_G)$ for $t$ steps from $\varphi_0=\varphi_R^\ast$. Then the measured gap is
\[
M_N(t)=\mathcal L_N(\varphi_t)-\mathcal L_R(\varphi_R^\ast)=(\mathcal L_N^\ast-\mathcal L_R^\ast)+E_N(t),\qquad
E_N(t)=\tfrac12\sum_i \lambda_i(1-\eta\lambda_i)^{2t}\,(q_i^\top d_N)^2,
\]
where $(\lambda_i,q_i)$ are eigenpairs of $H_N$ and $d_N=\varphi_R^\ast-\varphi_N^\ast$. $E_N(t)\ge0$ is non-increasing in $t$, and $E_N(t)\ge\tfrac12(1-\eta\varepsilon)^{2t}\sum_{\lambda_i\le\varepsilon}\lambda_i(q_i^\top d_N)^2$ for any $0<\varepsilon\le1/\eta$.
\end{proposition}
\begin{corollary}[Misranking]\label{cor:rank}
Let $N_1$ be equivalent to the real wiring ($\mathcal L_{N_1}^\ast=\mathcal L_R^\ast$) and $N_2$ genuinely worse by $g>0$. After $t$ steps, $N_1$ is ranked below $N_2$ whenever $E_{N_1}(t)>g+E_{N_2}(t)$. This happens when $N_1$'s optimum lies far from $\varphi_R^\ast$ along soft directions of $H$.
\end{corollary}
P1 is the case $t=0$. Longer budgets shrink the bias, but slowly along soft directions, which are common in circuit models \citep{gutenkunst2007}. The quadratic model is a local idealisation; we test its qualitative predictions in \cref{sec:flyvis}.

\section{Validation on synthetic ground truth}\label{sec:synth}
We generate scores for two wirings over 200 Latin-hypercube \citep{mckay1979} $\theta\in[0,1]^6$ with 10 noisy seeds each, from a shared performance landscape plus a planted structural effect. We then apply exactly the analysis used for the circuits, with 200 replicates per scenario. The criterion keeps $\theta$ where the real score reaches 0.8; ``data'' restrict $\theta_1\ge0.4$.

\begin{table}[h]
\centering\small
\caption{Protocols on synthetic ground truth (mean and 95\% range over replicates).}\label{tab:synth}
\resizebox{\linewidth}{!}{%
\begin{tabular}{lcccccc}
\toprule
Scenario & True effect (all / data) & P1 & P2 & P3 & Key param.\ named\\
\midrule
S1 no effect & $0$ / $0$ & $+0.004$ {\scriptsize[$-.004$,$.013$]} & $+0.001$ {\scriptsize[$-.021$,$.022$]} & $0.000$ {\scriptsize[$-.021$,$.020$]} & 20\% (chance 17\%)\\
S2 constant & $+.080$ / $+.080$ & $+0.082$ {\scriptsize[$.076$,$.090$]} & $+0.079$ {\scriptsize[$.059$,$.100$]} & $+0.079$ {\scriptsize[$.058$,$.099$]} & 19\%\\
S3 only if $\theta_1<0.4$ & $-.015$ / $-.045$ & $+0.028$ {\scriptsize[$.020$,$.036$]} & $+0.023$ {\scriptsize[$.003$,$.042$]} & $-0.004$ {\scriptsize[$-.027$,$.018$]} & 100\%\\
\bottomrule
\end{tabular}}
\end{table}

All protocols recover the no-effect and constant-effect scenarios (\cref{tab:synth}). With a conditional effect, P1 reports the wrong sign, because the criterion keeps the region where the planted advantage lives. P2 is positive, which is correct for its estimand: the real wiring does have the better best case. P3 moves toward the data-region truth but is conservative, because best-of-five within the region sits near its boundary; we therefore read P3 as a direction, not a magnitude. The deciding parameter is identified in every replicate when one exists. When none exists it is named only at chance, so its index must be read with its size. Scenario S3 has the structure we find below for the Kenyon-cell wiring.

\section{Three fly circuits}\label{sec:circuits}
\subsection{Mushroom body with dopamine-gated plasticity}\label{sec:mb}
\paragraph{Model.} All weights are FlyWire v783 synapse counts \citep{dorkenwald2024,schlegel2024}: 685 projection neurons, 5{,}177 Kenyon cells (KC$\gamma$ 2{,}489, KC$\alpha\beta$ 1{,}771, KC\aab{} 917), 96 output neurons (MBONs) and 323 dopaminergic neurons (DANs; PAM 307 for reward, PPL1 16 for punishment). A KC's drive is its number of active projection-neuron claws \citep{gruntman2013}. APL-like top-$k$ inhibition \citep{lin2014} selects active KCs with a fixed share per subtype. MBONs integrate KC input and interact through signed MBON--MBON synapses. The unconditioned stimulus drives PAM or PPL1, with signed MBON$\to$DAN feedback. KC$\to$MBON synapses are depressed when the KC is active and dopamine reaches that MBON's compartment through the connectome DAN$\to$MBON map, and each MBON's valence follows the dopamine family that dominates its input \citep{aso2014,li2020}. Sixty odours are learned in sequence with random valence; the score is final retention accuracy. Seven unmeasured parameters are swept (\cref{tab:mbparams}); 200 $\theta$ $\times$ 10 seeds, with 10 instances per null family.

\paragraph{Nulls.} KC identities shuffled; KC identities shuffled within subtype; degree-preserving KC$\to$MBON rewiring; degree-preserving PN$\to$KC rewiring; DAN$\to$MBON map rewired across reward and punishment families.

\begin{figure}[t]
\centering
\includegraphics[width=\linewidth]{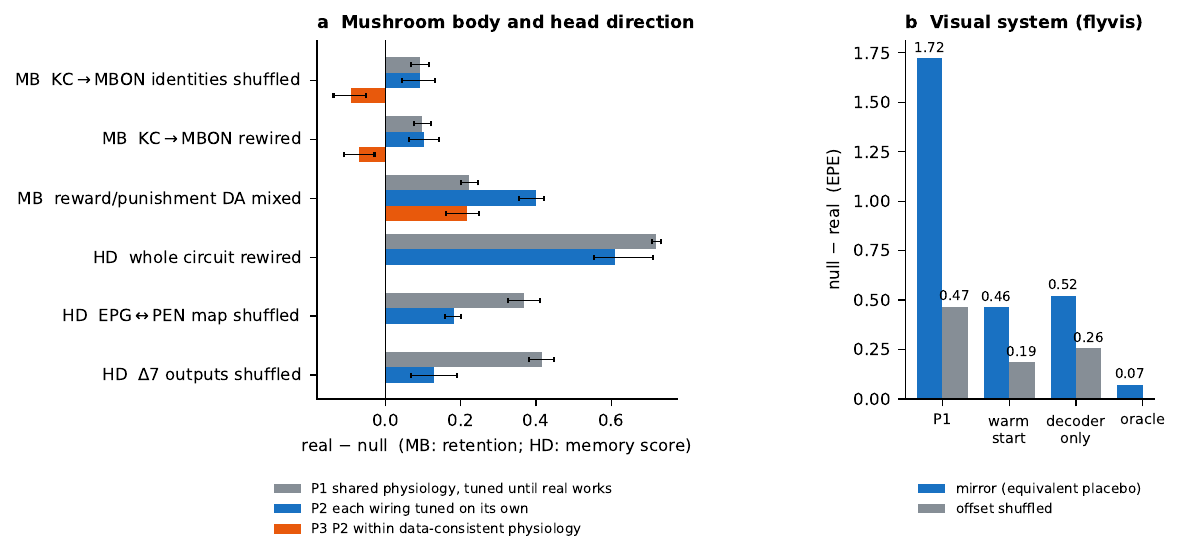}
\caption{\textbf{The comparison protocol decides fragile claims.} \textbf{a}, Real minus null under P1--P3 for six claims in the mushroom body (retention accuracy) and head direction (memory score); bars are bootstrap 95\% intervals over parameter sets. P3 exists only where independent data constrain an unmeasured parameter. The KC$\to$MBON claim reverses sign under P3; the other claims keep their sign. \textbf{b}, flyvis: null minus real validation EPE (median over members) for an exactly equivalent mirrored connectome and a spatially shuffled connectome, under P1, 2{,}000-iteration warm-start retraining, 2{,}000-iteration decoder-only retraining, and the exact oracle for the mirror.}
\label{fig:protocols}
\end{figure}

\paragraph{Results.} \emph{Dopamine family segregation is robust.} Real is better in 51\% of all $\theta$ and in 88\% of plausible $\theta$ (95 of 200, learning accuracy $\ge0.8$), with no $\theta$ favouring the null. All three protocols agree (P1 $+0.22$, P2 $+0.40$, P3 $+0.22$; \cref{fig:protocols}a). \emph{PN$\to$KC wiring and within-subtype KC identity make no clear difference} in 99\% and 97\% of $\theta$, consistent with largely random PN$\to$KC convergence \citep{caron2013}, though with some structure \citep{zheng2022}. This concerns our retention task only: biased PN$\to$KC sampling shapes odour categorisation by $\alpha\beta$ KCs \citep{chan2026}, which our task does not test.

\begin{figure}[t]
\centering
\includegraphics[width=\linewidth]{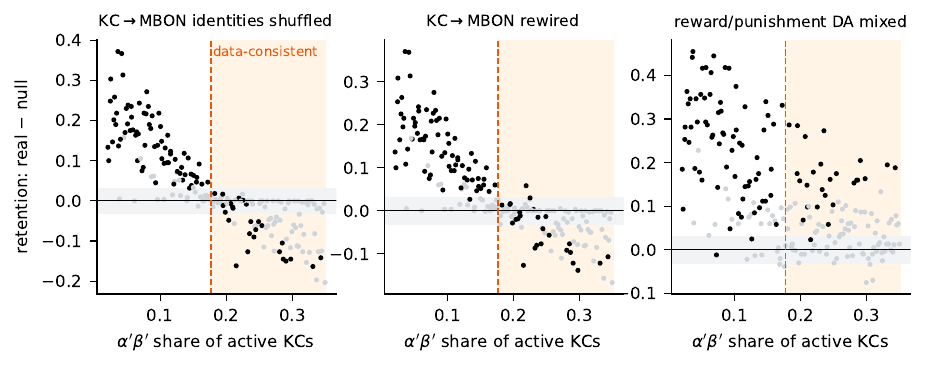}
\caption{\textbf{One unmeasured parameter decides the KC$\to$MBON claim.} Real minus null retention for each parameter set against the \aab{} share of active KCs. Black: the real model learns (plausible). Shaded: data-consistent region (\aab{} share above its 17.7\% share of cells). Grey band: $|\Delta|<0.03$. The crossing just above the 17.7\% reference point follows from the nulls mixing outputs across subtypes (see text). The dopamine-segregation claim (right) stays positive across the range.}
\label{fig:mb}
\end{figure}

\emph{The KC$\to$MBON advantage against cross-subtype nulls is decided by which KC subtype is active} (\cref{fig:mb}). Spearman $\rho=-0.88$ (identity shuffle) and $-0.87$ (rewiring); Sobol first-order index 0.88--0.89 and total 0.95 (surrogate $R^2$ 0.75--0.77). This dependence has a simple source, and its zero crossing is fixed by construction. Both nulls mix KC outputs across subtypes, so which subtype is active no longer matters to them: null retention is independent of the \aab{} share ($\rho=0.13$ and $0.06$), whereas real retention falls with it ($\rho=-0.60$). When each subtype's share of active KCs equals its share of all KCs (an \aab{} share of 0.177), the active cells are, by subtype, an unbiased sample of all KCs, so to first order the subtype organisation of outputs stops mattering and only finer, within-subtype structure remains. The value 0.177 is therefore a reference point fixed by construction, not an exact zero. Near it ($\pm0.025$, 31 $\theta$) $\Delta$ is small but positive, $+0.011$ [0.003, 0.020] and $+0.014$ [0.005, 0.024], about the size of the within-subtype effect there ($+0.009$); a running mean crosses zero at 0.185 and 0.20, and linear fits, which ignore curvature, at 0.22 and 0.23. The sign on either side reflects how useful each subtype's output wiring is for this task in this model: \aab{} outputs help less than average. Below 0.177, real is better in 39--42\% of $\theta$ (mean $\Delta=+0.12$) and no $\theta$ favours the null; above it, no $\theta$ favours real and 9--16\% favour the null. Shuffling KC identities \emph{within} subtype, which keeps the subtype composition of outputs, gives small effects with no dependence on \aab{} share ($\rho=0.06$; P1 $+0.019$, P2 $+0.002$, P3 $+0.038$). The KC$\to$MBON result is therefore a claim about subtype-level output organisation, not about individual KC wiring. 

\emph{Tuning until the model works selects the regime that favours the real wiring.} \aab{} KCs are reported to be the most excitable subtype \citep{inada2017}, so their share of active cells should exceed their share of cells, which places the system above 0.177; the verdict turns significantly negative only above about 0.2 (\cref{sec:measure}). This constraint is qualitative, so we vary the threshold from 0.10 to 0.25 (\cref{tab:p3thr}): at no threshold is the real advantage over cross-subtype nulls significantly positive. The behavioural criterion selects the other side, because real-model learning falls with \aab{} share ($\rho=-0.64$): only 26 of the 105 $\theta$ above 0.177 reach the learning criterion. Tuning until the model learns therefore selects the low-\aab{} regime, where the real wiring looks better. P1 $+0.091$ [0.068, 0.115], P2 $+0.093$ [0.045, 0.131], P3 $-0.092$ [$-0.138$, $-0.052$] for the identity shuffle, and $+0.098$, $+0.103$, $-0.069$ [$-0.111$, $-0.029$] for rewiring. These signs do not hinge on the random-search budget or on $K$: across 50 to 400 parameter sets (random subsets; main and held-out sweeps pooled for 400) and $K$ from 3 to 20, P2 is positive and P3 negative in at least 99\% of subsets for both nulls, and with $K=1$ in at least 97\% and 87\%. That the real model learns poorly in the data-consistent regime is itself a warning: either the model misses something that lets \aab{}-rich coding work, or the constraint is wrong; P3 cannot tell these apart. 

\begin{figure}[t]
\centering
\includegraphics[width=\linewidth]{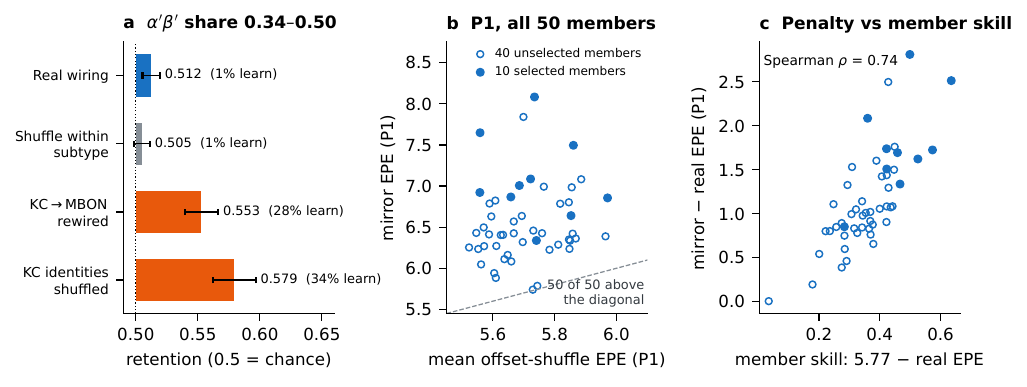}
\caption{\textbf{Key evidence.} \textbf{a}, Mushroom body at the \aab{} shares implied by imaging (0.34--0.50; 120 new $\theta$, 5 seeds): retention with bootstrap 95\% intervals, and in brackets the fraction of $\theta$ meeting the learning criterion. The real wiring is near chance and almost never learns; the cross-subtype nulls do better. \textbf{b}, flyvis under P1 for all 50 pretrained members: mirror EPE against the mean of three offset shuffles. Every member lies above the diagonal, so P1 penalises the mirror, which has a known near-real solution, more than a wiring with destroyed filter shapes. Filled: the ten members used elsewhere in the paper. \textbf{c}, The mirror penalty grows with how well the member learned the task (5.77 is the zero-flow baseline).}
\label{fig:evidence}
\end{figure}

\emph{Imaging data place the system where the real-wiring model fails} (\cref{fig:evidence}a). Subtype-resolved calcium imaging now constrains the share: Chan et al.\ \citep{chan2026} report average per-odour coding levels of about 10\% ($\alpha\beta$), 30\% (\aab{}) and 7\% ($\gamma$) KCs, which with connectome cell counts give an \aab{} share of active KCs of about 0.44 (about 0.37 for food odours) and an overall coding level of 12\%, inside our sparsity range. A sweep over shares of 0.34--0.50, with predictions documented before the run (Appendix~\ref{app:anchored}), gives $\Delta=-0.067$ [$-0.081$, $-0.053$] for the identity shuffle and $-0.040$ [$-0.051$, $-0.030$] for rewiring. But there the real-wiring model meets the learning criterion in 1 of 120 $\theta$ and its retention is near chance (0.51), whereas the cross-subtype nulls learn in 34\% and 28\% of $\theta$. At the imaging-derived composition the comparison therefore exposes an inadequacy of the model, not a demonstrated deficit of the fly's wiring. The failure may reflect omitted circuit mechanisms (for example subtype-specific APL feedback, which is linked to \aab{} excitability \citep{inada2017}), an inaccurate readout of \aab{} compartments, or a mismatch between our learning task and the conditions of the imaging data. Tuning until the model works hides this, because it selects \aab{} shares that the data exclude. 

\emph{The reversal holds on fresh samples, in model variants and against spatial nulls.} A held-out run with predictions documented before it started (new Latin hypercube, seeds and null instances; 200 $\theta$ $\times$ 10 seeds) confirmed all four predictions: cross-subtype null retention is again flat in \aab{} share ($\rho=0.09$, $0.02$; real $-0.66$), the crossings fall at 0.22 and 0.23, P1 is $+0.120$ [0.094, 0.147] and P3 $-0.070$ [$-0.122$, $-0.034$] for the identity shuffle, and the within-subtype effect stays small (mean $\Delta=+0.012$) with no \aab{} dependence ($\rho=-0.13$; Appendix~\ref{app:holdout}). The pattern also holds, again against predictions documented before the run, in a purely feedforward variant (no MBON--MBON or MBON$\to$DAN coupling; P1 $+0.073$, P3 $-0.091$) and with graded instead of claw-count KC input (P1 $+0.058$, P3 $-0.064$). In a variant without subtype-specific excitability, where a global winner-take-all sets which KCs fire, only 2--8\% of active KCs are \aab{}, the real wiring wins under every protocol (P1 $+0.19$), and there is no reversal (Appendix~\ref{app:variants}). A model without subtype-specific excitability therefore places the system in the regime that favours the real wiring without any parameter being tuned to put it there. A spatially constrained rewiring, in which each KC may swap MBON partners only with one of its 20 nearest KCs by soma position (47\% of edges unchanged, against 22\% for unconstrained rewiring), gives the same pattern with a smaller reversal: P1 $+0.090$ [0.072, 0.109], P2 $+0.116$ [0.084, 0.148], P3 $-0.031$ [$-0.066$, $-0.001$], \aab{} share again dominant ($\rho=-0.86$; Appendix~\ref{app:spatial}).

\emph{Heterogeneous dopamine responses} (Appendix~\ref{app:mbsupp}). In the main model every DAN of a family receives the same drive, so a within-family DAN shuffle cannot matter. Adding the spread of drive across individual DANs as an eighth parameter, a within-family shuffle null becomes informative. Its verdict is decided by that spread ($\rho=+0.46$): no clear difference in 94\% of $\theta$ with near-homogeneous drive, and real better in 47\% with strongly heterogeneous drive. Family segregation and the \aab{}-driven KC$\to$MBON result are unchanged.

\subsection{Head-direction ring attractor}\label{sec:hd}
\paragraph{Model.} From FlyWire v783 we take 155 neurons: EPG (47), EPGt (4), $\Delta$7 (42; glutamatergic, inhibitory), PEN\_a (20), PEN\_b (22) and PEG (20) \citep{hulse2021}. Each EPG's angle is taken from a Laplacian eigenmap of its EPG$\leftrightarrow$PEN connectivity (leading eigenvalue pair 1.27/1.39; ring radius CV 0.13). As an independent check, EPG$\to\Delta$7$\to$EPG inhibition, which is not used to assign angles, is strongest at $\pm\pi$ and weakest near 0, the expected opposite-side profile. Dynamics follow a saturating rate model, $\tau_c\dot r=-r+\mathrm{clip}_{[0,1]}(G_E\hat W_Er-G_I\hat W_Ir+b+I+\eta)$, with each neuron's excitatory and inhibitory input normalised by its own input total. Six parameters are swept (\cref{tab:hdparams}); 400 $\theta\times10$ seeds. \emph{Memory}: a cosine input at 12 angles followed by darkness, scored by $\cos$(decoded $-$ cued angle), and 0 if the bump dies or spreads. \emph{Integration}: opposite-signed angular-velocity drive to left and right PENs, scored by $|\mathrm{corr}(\text{velocity},\text{bump displacement})|$. Each null is decoded with its own ring embedding, so a permuted ring is not penalised. Plausible $\theta$ are those where the real bump persists in darkness \citep{seelig2015,kim2017}.

\paragraph{Results.} The real circuit holds a bump in 42\% of $\theta$; the shuffled PEN map in 10\%, shuffled $\Delta$7 outputs in 5\%, and both rewirings in 0\% of 400 sets. Among plausible $\theta$, real beats both rewirings in 100\% of sets for memory and 92\% for integration, and no $\theta$ favours any null for memory. Under P2 the real circuit scores 0.88 (memory) and 0.95 (integration), against 0.70/0.60 (PEN map), 0.75/0.40 ($\Delta$7) and $\le0.29$/$\le0.02$ (rewirings). Integration against the two targeted shuffles is robust in direction but moderate in strength (real better in 46--49\% of plausible $\theta$; null better in 9\% and 3\%). The deciding factor is the excitation--inhibition balance. For the whole-circuit memory claim (surrogate $R^2=0.60$), $G_E$ and $G_I$ have total-order Sobol indices 0.55 and 0.50 but first-order indices of only 0.24 and 0.30, so they act mainly through their interaction; surrogates for the targeted shuffles fit poorly ($R^2\le0.31$) and we do not interpret them. Replacing the angle assignment with one from EPG$\leftrightarrow$EPG connectivity, which uses no PEN wiring (circular agreement 0.94), leaves every verdict unchanged in direction (Appendix~\ref{app:hdsupp}). Spatially constrained rewiring, which swaps a partner only with one of the 5 nearest neurons by soma position (any type, or same type only; about half of edges unchanged), still loses to the real circuit in 100\% of plausible $\theta$ for memory; for integration the real circuit is better in 56\% (any type) and 92\% (same type) of plausible $\theta$, and the null in at most 4\% (Appendix~\ref{app:spatial}). For memory, P1 inflates every effect relative to P2 (by 0.11, 0.19 and 0.29; \cref{fig:protocols}a), as \cref{prop:gauss} predicts for a selective criterion. For integration the order reverses (P2 $>$ P1), because the criterion selects on memory, not on integration. The integration score is $|\mathrm{corr}|$ between velocity and bump displacement, which would also reward integration in the wrong direction. Scoring instead the signed correlation, oriented by the anatomical left/right PEN drive (fixed in advance, not per $\theta$), leaves every integration verdict positive and enlarges the advantage over the PEN and $\Delta$7 shuffles (P2 $+0.69$ and $+0.66$, against $+0.35$ and $+0.55$), because about half of the shuffled circuits that integrate do so in the wrong direction; the real circuit integrates in the anatomical direction in 86\% of plausible cases (Appendix~\ref{app:hdsupp}).

\subsection{Trained visual system (flyvis)}\label{sec:flyvis}
\paragraph{Model.} We use the pretrained flyvis ensemble \citep{lappalainen2024}: 45{,}669 neurons of 65 types on a hexagonal lattice, trained for optic flow on Sintel \citep{butler2012} for about 250{,}000 iterations. The connectome fixes synapse counts per type pair and spatial offset; the 734 trained parameters are the unmeasured physiology. We use the 10 best members and the package's original validation split. The members were ranked by the original authors on this same validation loss, and Sintel's test set has no public ground truth, so the real network's validation error is optimistically selected. This favours the real wiring under P1 and inflates the oracle gap (which subtracts the selected real error), so it makes our mirror comparison conservative. Selection on the real error may also influence null-versus-null comparisons, so we repeated P1 on all 50 members, including the 40 that were not selected. It gives the same picture (\cref{fig:evidence}b): the mirror is worse than real in 50 of 50 members and penalised more than the mean offset shuffle in 50 of 50 (median $+17.5\%$ on the unselected members, against $+32\%$ on the selected ten), and all three shuffles lose in 49 of 50. The mirror penalty grows with how well the member learned the task (Spearman $\rho=0.74$ against $5.77$ minus the real EPE; $\rho=0.71$ for the mirror-minus-shuffle penalty; \cref{fig:evidence}c), as expected if P1 measures how well physiology tuned for the real wiring fits another wiring. Our recomputed validation EPE matches the stored values (mean absolute difference 0.07). A network carrying no information (zero-flow prediction) has EPE 5.77.

\paragraph{Nulls.} \emph{offset\_shuffle}: within every type pair, synapse counts are permuted across spatial offsets, destroying filter shape and keeping totals. \emph{type\_rewire}: degree-preserving swaps of target cell types. \emph{mirror}: every offset reflected, $(du,dv)\to(-du,-dv)$, a 180\textdegree{} rotation of the lattice. Physiology is shared per cell type and type pair, so the mirrored network with the same physiology computes the rotated version of the real network. The mirror therefore has an achievable solution whose error equals the real network's error on 180\textdegree-rotated validation videos, which we compute exactly; this \emph{oracle} is an upper bound on the mirror's best error, not necessarily its global optimum, because the data need not be rotation-symmetric.

\begin{figure}[t]
\centering
\includegraphics[width=0.85\linewidth]{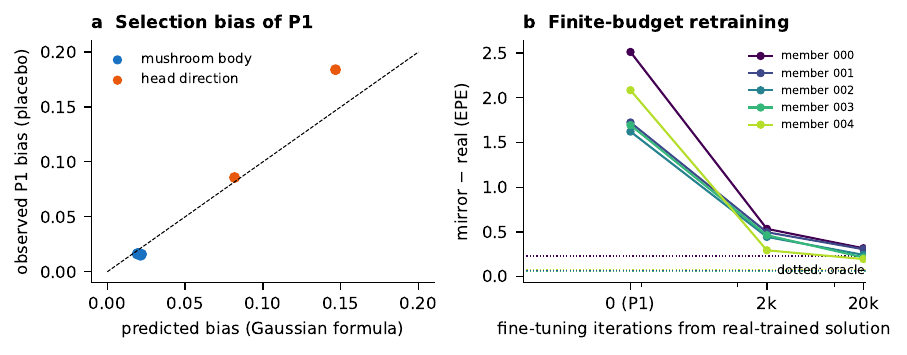}
\caption{\textbf{Theory against data.} \textbf{a}, Observed P1 bias in placebo comparisons (two instances of one null family, one labelled ``real'') against the Gaussian prediction of \cref{prop:gauss}, estimated from the same data. \textbf{b}, Mirror minus real EPE in flyvis against retraining budget from a real-trained solution; dotted lines mark each member's exact oracle gap.}
\label{fig:theory}
\end{figure}

\paragraph{Results.} Under P1 (physiology and decoder from the real-trained member), every null loses in 10 of 10 members: offset\_shuffle by 7\%, the mirror by 32\%, and type\_rewire diverges (median EPE 63). P1 thus penalises the mirror, which has a known near-real solution, more than the offset shuffle, which destroys filter shape. Budget-limited retraining does not repair this (\cref{fig:protocols}b, \cref{tab:flyvis}). With 2{,}000 iterations of warm-start retraining, the mirror stays near the zero-flow baseline (5.71 against 5.77) and below the offset shuffle (5.50); decoder-only retraining, where network parameters change by exactly zero, gives the same picture. The oracle shows the mirror can come within 0.07 EPE of real (paired median; range 0.05--0.23). A ten-fold longer budget (20{,}000 iterations, five members, each condition including real retrained identically) shrinks the mirror gap but does not correct the ranking (\cref{fig:theory}b, \cref{tab:flyvis}). The mirror's excess over real falls from a median of 1.72 (P1) to 0.46 (2k) and 0.24 (20k; range 0.20--0.32), above every member's oracle gap (median 0.07; range 0.05--0.23), with validation curves largely flat over the second half of training. The mirror remains ranked below the offset shuffle in 5 of 5 members (median excess 0.13; range 0.05--0.21; one-sided sign test $p=0.031$). Retraining from the converged solution does not improve the real network itself (median change $+0.08$), so the comparison is against the same procedure, not against an unchanged real network. For member 000 we extended warm-start retraining to 100{,}000 iterations for real, mirror and offset shuffle. The mirror reaches 5.29 and ends below the offset shuffle (5.43), but the real network degrades from 5.14 to 5.61, and the validation losses of the real and offset-shuffle runs rise in the second half of training while the mirror's keeps falling, so the final-checkpoint order reflects the degradation of the first two rather than a recovered mirror; by the lowest validation loss along each curve the mirror is again last. At this budget the ranking depends on which checkpoint is reported, so long warm-start fine-tuning with the default schedule is not a stable comparison either. This is consistent with \cref{prop:warm} and \cref{cor:rank}, although flyvis is trained with Adam rather than the gradient descent of the proposition: the mirror's known solution differs from the real one in the readout, which must reverse its mapping, and the shrinking but persistent excess matches the predicted slow decay. We have not measured the flyvis Hessian, so slow convergence along soft directions is a consistent explanation, not a demonstrated one.

\begin{table}[t]
\centering\small
\caption{flyvis validation EPE (median over members: 5 for P1, both warm-start budgets and oracle, 3 for decoder only; lower is better; zero-flow baseline 5.77).}\label{tab:flyvis}
\begin{tabular}{lccccc}
\toprule
Connectome & P1 & Warm, 2k & Warm, 20k & Decoder only, 2k & Oracle\\
\midrule
Real & 5.25 & 5.26 & 5.33 & 5.21 & 5.25\\
Mirror (equivalent placebo) & 7.01 & 5.71 & 5.60 & 5.72 & 5.36\\
Offset shuffle & 5.74 & 5.50 & 5.47 & 5.40 & ---\\
Type rewire & diverges & 5.77 & --- & --- & ---\\
\bottomrule
\end{tabular}
\end{table}

\section{Which measurements matter, and where the model fails}\label{sec:measure}
A sweep says which parameter a verdict depends on; an experimentalist needs to know which measurement would change the verdict. We simulate a measurement of one parameter as learning which fifth of its prior range it lies in (five equal-probability bins), and recompute the real-minus-null difference over plausible $\theta$ within each bin, with a bootstrap 95\% interval. A measurement can \emph{flip} a claim if some bins give a significantly positive and others a significantly negative difference; its \emph{spread} is the range of the bin means.

Of the 52 claim--parameter pairs (\cref{tab:measure}), two can flip a verdict, both the \aab{} share of active KCs, one for each KC$\to$MBON claim. For the identity shuffle, the real wiring wins below an \aab{} share of 0.15 ($+0.20$ [0.18, 0.22] and $+0.10$ [0.09, 0.12] in the two lowest bins), the difference is indistinguishable from zero between 0.15 and 0.22 ($+0.002$ [$-0.031$, 0.031]), and the null wins above 0.22 ($-0.06$ [$-0.09$, $-0.03$] and $-0.14$ [$-0.15$, $-0.13$]); rewiring gives the same pattern. These intervals are not corrected for multiple comparisons, and the five bins hold 36, 28, 14, 11 and 6 plausible $\theta$. The real-wiring advantage in the two lowest bins is clear-cut: every $\theta$ there favours the real wiring against both nulls (exact sign test $p<10^{-8}$, far below a Bonferroni level of $0.05/260$ for all $52\times5$ bin tests). The null advantage above 0.22 is less secure. Bootstrap percentile intervals at the Bonferroni level are [$-0.107$, $-0.012$] and [$-0.160$, $-0.112$] for the identity shuffle and [$-0.080$, 0.025] and [$-0.136$, $-0.080$] for rewiring, but percentile intervals this far in the tail are unreliable for so few Latin-hypercube points, and an exact sign test could not reach that level with 11 or 6 $\theta$ even if every sign agreed. We therefore treat the upper-bin reversal as exploratory rather than as controlling the familywise error rate. The held-out sweep reproduces the three lowest bins ($+0.23$, $+0.11$, $+0.002$) but has only 4 and 2 plausible $\theta$ in the upper bins; it supports the null advantage at high share in aggregate, through its P3 estimate and zero crossings (Appendix~\ref{app:holdout}). Within this model, an imaging experiment therefore needs only to tell whether \aab{} cells make up less than about 15\% or more than about 22\% of odour-responsive KCs. Subtype-resolved imaging \citep{chan2026} implies about 44\% (\cref{sec:mb}), which settles the model's verdict but also shows that the model fails to learn there, so the biological question needs a better model as well as the measurement.

For the other 50 claim--parameter pairs no single measurement changes the sign, even without correction, although this does not exclude effects too small to detect. Some set the size of the effect: the advantage of the real PEN loop falls from 0.72 to 0.22 across the range of the $\Delta$7 inhibitory gain, and the integration advantage of the real head-direction circuit from 0.79 to 0.26 across the excitatory gain. These are the claims where a measurement would sharpen a number rather than decide a direction.

\paragraph{Prior-weighted comparison.} P3 uses a hard threshold. A softer alternative averages $\Delta(\theta)$ over plausible $\theta$ weighted by a prior on the \aab{} share, $\E_{p}[\Delta\mid L_R\ge c]$, which is the comparison a measured distribution of that share would allow. With a Gaussian prior of standard deviation 0.04, the identity-shuffle difference is $+0.14$ [0.13, 0.16] for a prior mean of 0.10, $+0.036$ [0.019, 0.053] at 0.177, $-0.002$ [$-0.024$, 0.018] at 0.20, $-0.061$ [$-0.082$, $-0.041$] at 0.25 and $-0.109$ [$-0.126$, $-0.083$] at 0.30 (rewiring: $+0.14$, $+0.045$, $+0.012$, $-0.039$, $-0.081$). The verdict thus changes sign for a prior centred near 0.2. This is a sensitivity analysis, not a posterior: its estimand is the paired difference averaged over $\theta$, unlike the constrained best case of P3. The imaging estimate of 0.44 lies above this sign change, in the regime where, as shown in \cref{sec:mb}, the real-wiring model itself fails.

\begin{table}[t]\centering\small
\caption{Measurement priority. Pooled: P1 difference over plausible $\theta$. Most informative: parameter with the largest spread of bin means. Flip: whether pinning any single parameter to one fifth of its range can reverse the significant sign (uncorrected 95\% bootstrap intervals within each bin).}\label{tab:measure}
\begin{tabular}{lcccc}\toprule
Claim & pooled & most informative & spread & any flip\\\midrule
MB KC$\to$MBON vs identity shuffle & $+0.091$ & \aab{} share & 0.34 & yes (\aab{} share only)\\
MB KC$\to$MBON vs rewiring & $+0.098$ & \aab{} share & 0.31 & yes (\aab{} share only)\\
MB dopamine family segregation & $+0.223$ & \aab{} share & 0.14 & no\\
MB within-subtype shuffle & $+0.019$ & PN fraction & 0.06 & no\\
HD memory, whole-circuit rewiring & $+0.720$ & excitatory gain & 0.12 & no\\
HD memory, PEN loop shuffle & $+0.368$ & $\Delta$7 inhibitory gain & 0.51 & no\\
HD memory, $\Delta$7 output shuffle & $+0.415$ & excitatory gain & 0.36 & no\\
HD integration, whole-circuit rewiring & $+0.480$ & excitatory gain & 0.53 & no\\\bottomrule
\end{tabular}
\end{table}

\section{Placebo nulls and the size of the bias}\label{sec:placebo}
P1 shows a spurious advantage in every placebo family (\cref{tab:placebo}). Because the 90 ordered pairs reuse ten instances, we infer at the instance level: averaging over partners gives one value per instance labelled ``real'', and a $t$ interval over the ten instances excludes zero in five of six families (positive in 8--10 of 10 instances); for the identity shuffle the interval is [$-0.002$, 0.034], with 7 of 10 positive. The P2 placebo has no sign; its mean absolute value is of the same order as seed noise. The closed form of \cref{prop:gauss}, with covariances estimated from the placebo data, predicts the bias to within 40\% (5--40\% across families) and orders the families correctly (\cref{fig:theory}a; pairwise correlation between predicted and observed bias 0.60--0.90 across seed pairs). The bias is largest where the criterion is most selective. In head direction it accounts for a large part of the P1 effect against $\Delta$7 shuffling ($+0.18$ of $+0.42$). The KC$\to$MBON P1 effect ($+0.09$) is about six times its placebo bias ($+0.016$), yet it reverses under P3, so there the decisive factor is the unmeasured \aab{} share, not selection alone.

\begin{table}[h]
\centering\small
\caption{Placebo comparisons (10 instances, each with its own seed; means over 90 ordered instance pairs, inference over instances in the text).}\label{tab:placebo}
\begin{tabular}{llcccc}
\toprule
Circuit & Null family & $\theta$ kept & P1 bias & Predicted & $|$P2$|$\\
\midrule
Mushroom body & KC identities shuffled & 0.45 & $+0.016$ & $+0.022$ & 0.027\\
Mushroom body & KC$\to$MBON rewired & 0.44 & $+0.015$ & $+0.021$ & 0.019\\
Mushroom body & PN$\to$KC rewired & 0.46 & $+0.016$ & $+0.019$ & 0.008\\
Mushroom body & KC shuffled within subtype & 0.46 & $+0.017$ & $+0.020$ & 0.007\\
Head direction & EPG$\leftrightarrow$PEN map shuffled & 0.14 & $+0.086$ & $+0.082$ & 0.076\\
Head direction & $\Delta$7 outputs shuffled & 0.10 & $+0.184$ & $+0.147$ & 0.052\\
\bottomrule
\end{tabular}
\end{table}

\section{Discussion}
\textbf{A real-versus-null verdict at one parameter setting is weak evidence that wiring matters.} Across three circuits, the same procedure separates claims that hold under every protocol we test from claims that reverse, and names the quantity that decides each fragile claim. Observed performance depends jointly on topology, physiology, initialisation and optimisation budget, so a better score for the real wiring does not by itself identify which of these produced it. A structure--function claim needs three checks: a fair comparison (P2 rather than P1), physiology consistent with data (P3), and a model that still performs the task in that physiology. The mushroom body fails the third check at the imaging-derived subtype composition, which is itself a finding the first two protocols cannot reveal.

\paragraph{Recommendations.}
\begin{enumerate}[leftmargin=*,itemsep=1pt]
\item Report verdicts as fractions of a sampled space of unmeasured physiology, with the deciding parameter and its sensitivity index.
\item Report P2 beside shared-parameter comparisons; treat a claim that holds only under P1 as unconfirmed. Where independent data constrain the deciding parameter, report P3.
\item Include placebo nulls. Their P1 value estimates the protocol's own bias.
\item For trained models, do not treat finite-budget retraining from a real-trained solution as a neutral control. Where a wiring with a known near-equivalent solution exists (a symmetry of the lattice), use it as an oracle-checked placebo.
\item Check that the real-wiring model still performs the task in the data-consistent physiology before reading P3 as a statement about the animal.
\item Prefer targeted nulls that break one feature. Global rewiring often destroys everything and cannot say which feature matters.
\end{enumerate}

\paragraph{Measurements that would settle contested claims.} Whether the mushroom-body KC$\to$MBON wiring beats cross-subtype nulls depends on the share of \aab{} KCs among odour-responsive KCs, and only on whether it is below about 15\% or above about 22\% (\cref{sec:measure}); subtype-resolved imaging \citep{chan2026} places it near 0.44, where our model fails to learn, so the next step is a model that learns at the estimated composition. Whether the within-family dopamine map matters depends on how heterogeneous single-DAN responses to the same reinforcer are \citep{jiang2021}.

\paragraph{Limitations.}
\begin{itemize}[leftmargin=*,itemsep=1pt]
\item \emph{Model class.} The mushroom-body and head-direction models are rate models with transmitter-based signs; spiking or more detailed biophysical versions may behave differently. Our claims concern comparison protocols: the propositions do not depend on the model class, and the protocol bias also appears in the trained flyvis model.
\item \emph{Spatial embedding of nulls.} Degree-preserving rewiring ignores neuron geometry and wiring cost, which are needed to reproduce connectome structure \citep{salova2025}. Several of our nulls keep spatial structure (within-subtype shuffles, offset shuffles within each filter's spatial support, and the mirror, which is a lattice symmetry), and we add spatially constrained rewiring of the mushroom body and head-direction circuits (Appendix~\ref{app:spatial}), which leaves every verdict unchanged in sign. Soma proximity is only a proxy for wiring cost: we do not model arbour geometry or synapse placement, and nulls that conserve wiring length exactly remain to be built.
\item \emph{Physiological flexibility.} We sweep a few global parameters per circuit; with per-neuron gains trained for each wiring, shuffled compass wiring can hold a heading, though not on a ring \citep{yad2026}. Our head-direction verdicts concern the bump under the parameters we vary.
\item \emph{Head-direction angles} come from connectivity, not protoglomerulus annotations; two independent connectivity-based assignments agree, but anatomical labels would be preferable.
\item \emph{flyvis from-scratch training.} We did not compare converged from-scratch training of null connectomes. A pilot that trained the real connectome from scratch for 20{,}000 iterations, with settings matching those reported in \citep{lappalainen2024} except one regulariser weight (0.1 rather than 0.01 above baseline), stayed at the zero-flow baseline (EPE 5.80$\to$5.77). A full 250{,}000-iteration run of the real connectome with the same pipeline reached 5.59 (from 5.80), worse than most published members (range 5.13--5.74; median 5.43 over the 40 unselected ones; 5.31 over the selected ten), so our from-scratch pipeline does not reproduce the original training; a converged from-scratch comparison of the nulls is left to future work. Our flyvis conclusions rest on P1, budget-limited retraining and the oracle, which needs no training. The residual oracle gap (median 0.07) is not yet explained; the ensemble was trained with flip and rotation augmentation, and imperfect equivariance of the trained solution is one candidate.
\item \emph{Statistics.} Per-$\theta$ verdicts are descriptive, without correction across $\theta$; protocol effects carry bootstrap intervals that resample parameter sets, keeping real and null paired within each set. A Latin hypercube is not an i.i.d.\ sample, so these intervals are approximate; the held-out and variant sweeps (Appendices~\ref{app:holdout} and~\ref{app:variants}) use independent hypercubes and serve as replicate-level checks. The ten-fold budget test uses five members and one offset-shuffle instance.
\item \emph{Exploratory analysis.} The mushroom-body result was found by exploration; we confirmed it on fresh parameter sets, seeds and null instances against four predictions documented before the run, all of which held (Appendix~\ref{app:holdout}). The prediction files were not registered with a third party; their file timestamps document the order of work but cannot verify it independently. The result also holds in feedforward and graded-input variants and disappears, as predicted, when subtype excitability is removed (Appendix~\ref{app:variants}). Only 11 held-out $\theta$ are both plausible and above 0.177, so the P3 estimate rests on few points. Why \aab{} outputs help less than average in this model is not explained; it may not hold in the fly.
\item \emph{Ranges and P3.} Parameter ranges are literature-informed modelling choices, and the robustness map is conditional on them. P3 is available for one parameter in one circuit, and is conservative (\cref{sec:synth}).
\item \emph{Theory.} \Cref{prop:gauss} is a Gaussian approximation and \cref{prop:warm} a local quadratic one; we use them for their qualitative and order-of-magnitude predictions.
\end{itemize}

\paragraph{Conclusion.} Connectome-constrained models can reveal robust structure--function relationships, but an advantage over rewired controls is not sufficient evidence on its own. Parameter selection, physiological uncertainty and the optimisation procedure can each change how a wiring comparison should be read. A credible structural claim needs a comparison matched to the question, constraints grounded in independent measurements, and a model that still performs the task under those constraints.

\paragraph{Code and data.} All models, sweeps, analyses, prediction files, result tables and the Colab notebooks for flyvis are provided as ancillary files with this submission.

\paragraph{Use of AI-assisted tools.} 
The author used generative AI assistants (Claude and ChatGPT) for coding, analysis support, derivation checking, and manuscript editing. The author independently directed the research, verified the results, and takes full responsibility for the content and conclusions.

\bibliographystyle{unsrtnat}
\bibliography{refs}

\appendix
\section{Proofs}\label{app:proofs}
\paragraph{Proposition~\ref{prop:exch}.} Write $A=\mathbf 1\{S_R\ge c\}$, $B=\mathbf 1\{S_N\ge c\}$. By exchangeability $\E[S_NA]=\E[S_RB]$, so $\E[(S_R-S_N)A]=\E[S_R(A-B)]$. Exchangeability also gives $\E[S_R(A-B)]=\E[S_N(B-A)]$, so averaging the two expressions, $\E[(S_R-S_N)A]=\tfrac12\E[(S_R-S_N)(A-B)]$. If $S_R>S_N$ then $A\ge B$; if $S_R<S_N$ then $A\le B$. Hence $(S_R-S_N)(A-B)\ge0$ pointwise, with strict inequality exactly when the two scores fall on opposite sides of $c$. Dividing by $\Pr(A=1)$ gives the claim. \qed

\paragraph{Proposition~\ref{prop:gauss}.} For jointly Gaussian $(X,L)$, $\E[X\mid L]=\E X+\frac{\Cov(X,L)}{\sigma_L^2}(L-\mu_L)$. The tower property gives $\E[X\mid L\ge c]=\E X+\frac{\Cov(X,L)}{\sigma_L^2}\E[L-\mu_L\mid L\ge c]$, and for a Gaussian $\E[L-\mu_L\mid L\ge c]=\sigma_L\phi(\alpha)/(1-\Phi(\alpha))$ with $\alpha=(c-\mu_L)/\sigma_L$ and $1-\Phi(\alpha)=p$. If $L=S_R$ and $(S_R,S_N)$ are exchangeable with variance $\sigma^2$ and correlation $\rho$, then $\Cov(S_R-S_N,S_R)=\sigma^2-\rho\sigma^2$ and $\sigma_L=\sigma$, giving $\sigma(1-\rho)\lambda(\alpha)$. \qed

\paragraph{Proposition~\ref{prop:warm}.} For the quadratic loss, $\varphi_{t+1}-\varphi_N^\ast=(I-\eta H_N)(\varphi_t-\varphi_N^\ast)$, so $\varphi_t-\varphi_N^\ast=(I-\eta H_N)^t d_N$ and $\mathcal L_N(\varphi_t)-\mathcal L_N^\ast=\tfrac12 d_N^\top(I-\eta H_N)^tH_N(I-\eta H_N)^td_N$, since these matrices commute. Diagonalising $H_N$ gives the sum. For $0<\eta\le1/\lambda_{\max}$, every factor $(1-\eta\lambda_i)^{2t}\in[0,1]$ is non-increasing in $t$, so $E_N(t)$ is non-negative and non-increasing. Dropping the terms with $\lambda_i>\varepsilon$ and using $(1-\eta\lambda_i)\ge(1-\eta\varepsilon)$ for $\lambda_i\le\varepsilon$ gives the bound. The real wiring starts at its optimum, so its excess is zero and $M_N(t)$ follows. The corollary is the comparison $M_{N_1}(t)>M_{N_2}(t)$. \qed

\section{Mushroom-body details and supplementary results}\label{app:mbsupp}
\begin{table}[h]
\centering\small
\caption{Unmeasured mushroom-body parameters (Latin hypercube; log-uniform where marked).}\label{tab:mbparams}
\begin{tabular}{lll}
\toprule
Parameter & Range & Meaning\\
\midrule
\aab{} share & 0.02--0.35 & fraction of active KCs that are \aab\\
KC sparsity & 0.03--0.15 (log) & fraction of KCs active per odour \citep{honegger2011}\\
Plasticity rate $\eta$ & 0.1--0.9 (log) & dopamine-gated depression rate\\
$G_{MM}$ & 0--1 & MBON--MBON coupling gain\\
$G_{FB}$ & 0--1 & MBON$\to$DAN feedback gain\\
PN fraction & 0.05--0.3 (log) & fraction of PNs activated by an odour\\
Trial noise & 0--0.3 & trial-to-trial input dropout\\
DAN heterogeneity $\sigma$ & 0--1.5 & spread of US drive across DANs (supplementary sweep only)\\
\bottomrule
\end{tabular}
\end{table}
Remaining $\gamma$ and $\alpha\beta$ shares are split by cell counts. KC$\to$MBON weights are row-normalised per MBON; DAN$\to$MBON input is row-normalised. In the heterogeneity sweep, each DAN's US drive is multiplied by a fixed log-normal gain with mean 1 and spread $\sigma$ (200 $\theta\times4$ seeds). Within-family DA shuffle: no clear difference in 94\% of $\theta$ for $\sigma<0.5$ (mean $\Delta$ $+0.008$), real better in 21\% for $0.5\le\sigma<1$ and 47\% for $\sigma\ge1$ (mean $\Delta$ $+0.044$); no $\theta$ favours the null. Cross-family segregation: real better in 98\% of plausible $\theta$, advantage increasing with $\sigma$. KC$\to$MBON: $\rho(\sigma,\Delta)=-0.05$.

\begin{table}[h]
\centering\small
\caption{P3 against the data threshold on the \aab{} share (bootstrap 95\% intervals over parameter sets).}\label{tab:p3thr}
\begin{tabular}{lcccc}
\toprule
Threshold & $\theta$ in region & KC identities shuffled & KC$\to$MBON rewired & DA families mixed\\
\midrule
$\ge0.10$ & 152 & $-0.014$ {\scriptsize[$-.058$,$.027$]} & $+0.008$ {\scriptsize[$-.041$,$.053$]} & $+0.293$ {\scriptsize[$.258$,$.310$]}\\
$\ge0.13$ & 134 & $-0.035$ {\scriptsize[$-.085$,$.008$]} & $-0.012$ {\scriptsize[$-.065$,$.035$]} & $+0.273$ {\scriptsize[$.228$,$.291$]}\\
$\ge0.15$ & 122 & $-0.046$ {\scriptsize[$-.108$,$-.002$]} & $-0.023$ {\scriptsize[$-.082$,$.020$]} & $+0.262$ {\scriptsize[$.201$,$.285$]}\\
$\ge0.177$ & 105 & $-0.092$ {\scriptsize[$-.135$,$-.054$]} & $-0.069$ {\scriptsize[$-.113$,$-.025$]} & $+0.218$ {\scriptsize[$.166$,$.249$]}\\
$\ge0.20$ & 91 & $-0.123$ {\scriptsize[$-.152$,$-.085$]} & $-0.100$ {\scriptsize[$-.124$,$-.062$]} & $+0.187$ {\scriptsize[$.144$,$.224$]}\\
$\ge0.25$ & 61 & $-0.146$ {\scriptsize[$-.157$,$-.112$]} & $-0.114$ {\scriptsize[$-.128$,$-.087$]} & $+0.158$ {\scriptsize[$.117$,$.185$]}\\
\bottomrule
\end{tabular}
\end{table}

\section{Head-direction details and supplementary results}\label{app:hdsupp}
\begin{table}[h]
\centering\small
\caption{Unmeasured head-direction parameters.}\label{tab:hdparams}
\begin{tabular}{lll}
\toprule
Parameter & Range & Meaning\\
\midrule
$G_E$ & 0.5--8 (log) & summed excitatory input gain per neuron\\
$G_I$ & 0.5--8 (log) & summed $\Delta$7 inhibitory input gain per neuron\\
$b$ & $-0.5$--0.5 & tonic drive\\
$\tau_{\Delta7}/\tau_{EPG}$ & 0.5--5 (log) & $\Delta$7 time constant\\
$\tau_{PEN,PEG}/\tau_{EPG}$ & 0.5--5 (log) & PEN/PEG time constant\\
Noise & 0--0.1 & input noise\\
\bottomrule
\end{tabular}
\end{table}
Nulls: degree-preserving rewiring within every (post-type, pre-type) block; degree-preserving rewiring of the whole circuit; EPG$\to$PEN and PEN$\to$EPG maps shuffled by permuting PEN identities; $\Delta$7 outputs shuffled by permuting $\Delta$7 identities. With the alternative EPG$\leftrightarrow$EPG angle assignment, plausible $\theta$: 168 (vs 169); memory, real better than block/whole rewiring 100\%/100\% and than PEN/$\Delta$7 shuffles 79\%/80\%; integration 95\%/95\% and 57\%/58\%; null better at most 8\%.

\paragraph{Signed integration.} Rerunning the head-direction sweep (same $\theta$, seeds and null instances; memory and unsigned integration scores reproduce exactly) and storing the signed correlation, oriented so that positive means the bump moves in the direction set by the anatomical left/right PEN drive, gives, for whole-circuit rewiring, P1 $+0.354$ [0.287, 0.417] and P2 $+0.939$ [0.889, 0.969]; PEN loop shuffle P1 $+0.357$ [0.293, 0.422], P2 $+0.689$ [0.605, 0.757]; $\Delta$7 output shuffle P1 $+0.335$ [0.274, 0.396], P2 $+0.664$ [0.616, 0.707]. Among plausible $\theta$, the real circuit integrates ($|\mathrm{corr}|>0.3$) in 71\% of runs, 86\% of them in the anatomical direction; the PEN and $\Delta$7 shuffles integrate in 35\% and 26\% of runs, with the direction close to chance (51\% and 55\%); the rewirings almost never integrate.

\section{Spatially constrained nulls}\label{app:spatial}
Unconstrained degree-preserving rewiring can connect neurons that are far apart. We therefore rerun the main rewiring nulls with a spatial constraint. A double-edge swap $(a_1,b_1),(a_2,b_2)\to(a_1,b_2),(a_2,b_1)$ is proposed only if presynaptic $b_2$ is one of the $k$ nearest neighbours of $b_1$ by FlyWire soma position, so every neuron keeps its in- and out-degree and inputs move only between nearby cells. Mushroom body: KC$\to$MBON, $k=20$ (nearest KCs share a subtype 55\% of the time against a 38\% base rate). Head direction: whole circuit, $k=5$ among all neurons (\emph{any type}) or among neurons of the same type (\emph{same type}); soma-neighbouring EPGs have similar angles (mean difference 0.61 rad against 1.57 for random pairs). Five swap attempts per edge; accepted swaps 1.6 per edge (MB) and 1.4--1.6 (HD). Same $\theta$, seeds and real-model runs as the main sweeps (real scores identical).

\begin{table}[h]
\centering\small
\caption{Spatially constrained versus unconstrained rewiring (bootstrap 95\% intervals over $\theta$). Plausible: MB, real model learns ($n=95$); HD, real bump persists ($n=169$).}\label{tab:spatial}
\resizebox{\linewidth}{!}{%
\begin{tabular}{llccccc}
\toprule
Circuit, metric & Null & Edges kept & P1 & P2 & P3 & Plausible: real / null better\\
\midrule
MB, retention & KC$\to$MBON rewired & 22\% & $+0.098$ & $+0.103$ & $-0.069$ {\scriptsize[$-.111$,$-.029$]} & 42\% / 7\%\\
 & KC$\to$MBON spatial ($k=20$) & 47\% & $+0.090$ {\scriptsize[.072,.109]} & $+0.116$ {\scriptsize[.084,.148]} & $-0.031$ {\scriptsize[$-.066$,$-.001$]} & 39\% / 1\%\\
\midrule
HD, memory & whole circuit spatial & 50\% & $+0.49$ {\scriptsize[.45,.53]} & $+0.32$ {\scriptsize[.30,.34]} & -- & 100\% / 0\%\\
 & same-type spatial & 55\% & $+0.70$ {\scriptsize[.69,.72]} & $+0.61$ {\scriptsize[.59,.66]} & -- & 100\% / 0\%\\
HD, integration & whole circuit spatial & 50\% & $+0.31$ {\scriptsize[.26,.37]} & $+0.45$ {\scriptsize[.40,.49]} & -- & 56\% / 4\%\\
 & same-type spatial & 55\% & $+0.48$ {\scriptsize[.44,.52]} & $+0.93$ {\scriptsize[.88,.96]} & -- & 92\% / 0\%\\
\bottomrule
\end{tabular}}
\end{table}

For the MB, P3 against the \aab{} threshold with the spatial null is $+0.035$ [$-0.001$, 0.063] at $\ge0.10$, $+0.006$ [$-0.043$, 0.034] at $\ge0.15$, $-0.031$ [$-0.066$, $-0.001$] at $\ge0.177$, $-0.056$ [$-0.076$, $-0.029$] at $\ge0.20$ and $-0.066$ [$-0.081$, $-0.038$] at $\ge0.25$. As with the unconstrained nulls, the real advantage is never significantly positive in the data-consistent region. Per-$\theta$ differences with the spatial and unconstrained KC$\to$MBON nulls are almost perfectly rank-correlated (Spearman 0.99), so the \aab{} dependence does not come from spatially implausible rewiring. The reversal is smaller, consistent with the spatial null keeping twice as many real edges and moving inputs preferentially among same-subtype KCs. For the HD circuit, ring topography is required whether or not rewiring respects soma position. Integration against whole-circuit spatial rewiring is the weakest HD claim: robust in direction, moderate in strength.

\section{Held-out confirmation of the mushroom-body result}\label{app:holdout}
The mushroom-body analysis in \cref{sec:mb} was exploratory. Before running a confirmation sweep, we wrote down four predictions and the design (ancillary file \texttt{predictions/holdout.md}). The file is timestamped 3 November 2026, 13:09 EST, about a minute before the sweep started, and was not modified afterwards. It was not registered with a third party; these self-recorded timestamps document the order of work but cannot verify it independently, so we call the predictions documented in advance, not preregistered. The design used a new Latin hypercube (seed 1, 200 $\theta$), new simulation seeds (10--19, ten per $\theta$) and new null instances (pool seed 999) for the identity shuffle, the unconstrained KC$\to$MBON rewiring and the within-subtype shuffle. Metric, plausibility criterion and protocols are unchanged. \Cref{tab:holdout} gives the outcome; all four predictions held.

\begin{table}[h]\centering\small
\caption{Held-out confirmation (200 new $\theta$, seeds 10--19, new null instances). 79 $\theta$ are plausible, 105 lie above 0.177 and 11 are both.}\label{tab:holdout}
\begin{tabular}{p{6.3cm}p{6.4cm}c}\toprule
Prediction & Outcome & Held\\\midrule
H1: cross-subtype null retention flat in \aab{} share ($|\rho|<0.2$); real falls ($\rho<-0.4$) & null $\rho=0.09$ (shuffle), $0.02$ (rewiring); real $\rho=-0.66$ & yes\\
H2: zero crossing of $\Delta$ in [0.13, 0.23] & 0.218 (shuffle), 0.228 (rewiring) & yes\\
H3: identity shuffle P1 $>0$ and P3 $<0$, CIs excluding 0 & P1 $+0.120$ [0.094, 0.147]; P3 $-0.070$ [$-0.122$, $-0.034$] & yes\\
H4: within-subtype $|\text{mean }\Delta|<0.03$, $|\rho|<0.2$ & mean $\Delta=+0.012$, $\rho=-0.13$ & yes\\\bottomrule
\end{tabular}
\end{table}

For the rewiring null, P1 is $+0.124$ [0.098, 0.150], P2 $+0.119$ [0.097, 0.149] and P3 $-0.045$ [$-0.095$, $-0.013$]; for the identity shuffle P2 is $+0.113$ [0.078, 0.144]. The within-subtype shuffle gives P1 $+0.018$ [0.013, 0.024], P2 $+0.002$ [$-0.008$, 0.017] and P3 $+0.021$ [0.004, 0.045]: small, and positive under P1 and P3. The linear-fit crossings again lie somewhat above the base rate of 0.177, as in the exploratory sweep (0.22, 0.23). The P3 estimates rest on only 11 $\theta$, so their intervals are wide.

\section{Model-family robustness of the mushroom-body result}\label{app:variants}
We ran three variants of the mushroom-body model, with predictions documented before the runs (\texttt{predictions/variants.md}, timestamped 3 November 2026, 20:06 EST, under a minute before the first sweep started; not registered with a third party): a new Latin hypercube (seed 2, 120 $\theta$), seeds 20--24 and new null instances (pool seed 2024). \emph{ff} removes MBON--MBON and MBON$\to$DAN coupling, leaving a purely feedforward PN$\to$KC$\to$MBON circuit. \emph{norm} drives each KC by the fraction of its PN synapses that are active instead of by the number of active PN partners. \emph{noshare} drops subtype-specific excitability, so a global top-$k$ decides which KCs fire. We predicted the reversal in \emph{ff} and \emph{norm} (V1: flat null, crossing in [0.13, 0.25], P1 $>0$ and P3 $<0$ with CIs excluding 0), and for \emph{noshare} no dependence on the (now unused) \aab{} parameter, with the sign of $\Delta$ set by the realised \aab{} share (V2). All predictions held (\cref{tab:variants}).

\begin{table}[h]\centering\small
\caption{Model variants, identity-shuffle null (rewiring null in brackets).}\label{tab:variants}
\begin{tabular}{lccccc}\toprule
Variant & null $\rho$ & crossing & P1 & P3 & held\\\midrule
ff & 0.07 (0.04) & 0.21 (0.22) & $+0.073$ [0.037, 0.110] & $-0.091$ [$-0.131$, $-0.043$] & V1 yes\\
norm & 0.08 (0.02) & 0.21 (0.23) & $+0.058$ [0.033, 0.084] & $-0.064$ [$-0.095$, $-0.036$] & V1 yes\\
noshare & $\Delta$: 0.03 & none & $+0.187$ [0.171, 0.204] & -- & V2 yes\\\bottomrule
\end{tabular}
\end{table}

In \emph{noshare} the realised \aab{} share among active KCs is 4.6\% (median; range 1.8--8.2\%), far below their 17.7\% share of cells and opposite to the reported high excitability of \aab{} KCs \citep{inada2017}. P3 is not defined there because the \aab{} parameter is not used. The within-subtype shuffle again gives small effects in every variant ($|\text{mean }\Delta|\le0.015$).

\section{Data-anchored sweep of the \aab{} share}\label{app:anchored}
Chan et al.\ \citep{chan2026} define the coding level as the fraction of KCs responding to an odour (activity at least four standard deviations above baseline in at least two trials) and report average values of about 10\%, 30\% and 7\% for $\alpha\beta$, \aab{} and $\gamma$ KCs (five flies per type, 12 monomolecular odours, type-specific Gal4 drivers). With FlyWire counts (1{,}771, 917 and 2{,}489) the \aab{} share of active KCs is $917\cdot0.30/(917\cdot0.30+1771\cdot0.10+2489\cdot0.07)=0.44$. This lies above our original range (0.02--0.35), so before running we wrote down a design and predictions (\texttt{predictions/anchored.md}, timestamped 4 November 2026, 02:55 EST, under a minute before the sweep started; not registered with a third party): share uniform on [0.34, 0.50], other parameters as before, new hypercube (seed 3), 120 $\theta$, seeds 30--34, new null instances; prediction A1, mean $\Delta<0$ over all $\theta$ for both cross-subtype nulls with intervals excluding zero, and over plausible $\theta$ if at least ten are plausible; A2, within-subtype $|\text{mean }\Delta|<0.03$. A1 held over all $\theta$ ($-0.067$ [$-0.081$, $-0.053$]; $-0.040$ [$-0.051$, $-0.030$]; the null is better in 59 and 43 of 120 $\theta$, real in none), and A2 held ($+0.007$). The plausible part of A1 could not be tested: only 1 of 120 $\theta$ is plausible, because real retention is 0.512 [0.505, 0.519] and learning accuracy 0.63; the identity shuffle reaches 0.579 and 0.75, rewiring 0.553 and 0.73, and the within-subtype shuffle matches real (0.505, 0.63). The cross-subtype nulls meet the learning criterion in 34\% and 28\% of $\theta$. The original sweep shows the same above a share of 0.30 (real retention 0.53, nulls 0.60 and 0.58; 2 of 30 $\theta$ plausible). The imaging values are approximate averages, so the share is an estimate, not a per-odour measurement of the model parameter. It would fall to 0.22, the upper end of the range where real and null tie (\cref{sec:measure}), only if the $\alpha\beta$ coding level were 45\% instead of 10\%, the $\gamma$ level 32\% instead of 7\%, or the \aab{} level 11\% instead of 30\%.

\section{flyvis details}\label{app:flyvis}
flyvis 1.2.0 with the released 50-member ensemble (flow/0000). Pretrained configurations omit the split flag; we set the package's original train/validation split so that errors are comparable. Physiology transfer copies per-type resting potentials, time constants and per-type-pair synapse strengths, keeping each connectome's own synapse counts and signs. Warm-start and decoder-only retraining use Adam with the package's default schedule (learning rate stepped from $5\times10^{-5}$ to $5\times10^{-6}$, batch size 4); decoder-only runs zero the network learning rates. The oracle evaluates the real network and decoder on validation videos and flow targets rotated by 180\textdegree{} with the package's hexagonal rotation. Runs used one Colab GPU (A100 or L4); 2{,}000 iterations took about 3 minutes.

\end{document}